\documentclass[a4paper,fleqn]{cas-dc}

\usepackage[numbers]{natbib}
\usepackage{amsthm,amsfonts,amsmath,amssymb}
\usepackage[ruled,vlined]{algorithm2e}
\usepackage{pifont}
\usepackage{subfigure} 
\usepackage{graphicx}
\usepackage{psfrag}

\usepackage{xcolor}
\usepackage{array}
\usepackage{units}
\usepackage{mdframed}
\usepackage{colortbl}
\usepackage{setspace}
\usepackage{textcomp}
\usepackage{multirow}
\usepackage{enumitem}
\usepackage{auto-pst-pdf}
\usepackage{booktabs} 
\usepackage{steinmetz}
\usepackage{tabularray}

\makeatletter
\newcommand{\mypm}{\mathbin{\mathpalette\@mypm\relax}}
\newcommand{\@mypm}[2]{\ooalign{%
  \raisebox{.1\height}{$#1+$}\cr
  \smash{\raisebox{-.6\height}{$#1-$}}\cr}}
\makeatother
\usepackage[ruled,vlined]{algorithm2e}
\def\tsc#1{\csdef{#1}{\textsc{\lowercase{#1}}\xspace}}
\tsc{WGM}
\tsc{QE}
\tsc{EP}
\tsc{PMS}
\tsc{BEC}
\tsc{DE}

\begin{document}
\let\WriteBookmarks\relax
\def\floatpagepagefraction{1}
\def\textpagefraction{.001}

\shorttitle{A DRL Environment and Benchmark for Energy Arbitrage in Distribution Networks}

\shortauthors{Hou Shengren et~al.}

\title [mode = title]{RL-ADN: A High-Performance Deep Reinforcement Learning Environment for Optimal Energy Storage Systems Dispatch in Active Distribution Networks}                      
\tnotemark[1]

\tnotetext[1]{This publication is part of the project ALIGN4Energy (with project number NWA.1389.20.251) of the research programme NWA ORC 2020 which is (partly) financed by the Dutch Research Council (NWO), The Netherlands. This work is part of the DATALESs project (with project number 482.20.602) jointly financed by the Netherlands Organization for Scientific Research (NWO), and the National Natural Science Foundation of China (NSFC).}


%
\author[1]{Hou Shengren}[type=editor,
                        style=chinese,auid=000,
]

\credit{Conceptualization, Methodology, Software, Validation, Writing - original draft}

\affiliation[1]{organization={Department of Electrical Sustainable Energy, Delft University of Technology},
    addressline={Mekelweg 4}, 
    city={Delft},
    postcode={2628CD}, 
    country={The Netherlands}}

\author[1]{Gao Shuyi}[type=editor,
                        style=chinese,auid=000,
]
\credit{Writing - review \& editing}

\author[1]{Xia Weijie}[type=editor,
                        style=chinese,auid=000,
]
\credit{Writing - review \& editing}

\author[2]{Edgar Mauricio Salazar Duque}

\credit{review \& editing}

\author[1]{Peter Palensky}[%
 ]

\credit{Funding acquisition}

\author%
[1]
{Pedro P. Vergara}
\cormark[1]

\credit{Writing - review \& editing, Supervision, Funding acquisition}

\affiliation[2]{organization={Energy Systems Systems Group, Eindhoven University of Technology.},
    city={Eindhoven},
    postcode={5612 AE}, 
    country={The Netherlands}}

\cortext[cor1]{Corresponding author email: p.p.vergarabarrios@tudelft.nl}



\begin{abstract}
Deep Reinforcement Learning (DRL) presents a promising avenue for optimizing Energy Storage Systems (ESSs) dispatch in distribution networks. This paper introduces RL-ADN, an innovative open-source library specifically designed for solving the optimal ESSs dispatch in active distribution networks. RL-ADN offers unparalleled flexibility in modeling distribution networks, and ESSs, accommodating a wide range of research goals. A standout feature of RL-ADN is its data augmentation module, based on Gaussian Mixture Model and Copula (GMC) functions, which elevates the performance ceiling of DRL agents. Additionally, RL-ADN incorporates the Laurent power flow solver, significantly reducing the computational burden of power flow calculations during training without sacrificing accuracy. The effectiveness of RL-ADN is demonstrated using in different sizes of distribution networks, showing marked performance improvements in the adaptability of DRL algorithms for ESS dispatch tasks. This enhancement is particularly beneficial from the increased diversity of training scenarios. Furthermore, RL-ADN achieves a tenfold increase in computational efficiency during training, making it highly suitable for large-scale network applications. The library sets a new benchmark in DRL-based ESSs dispatch in distribution networks and it is poised to advance DRL applications in distribution network operations significantly. RL-ADN is available at: \url{https://github.com/ShengrenHou/RL-ADN} and \url{https://github.com/distributionnetworksTUDelft/RL-ADN}.

\end{abstract}


\begin{highlights}
\item Versatile Library: RL-ADN offers flexible DRL-based ESS dispatch in active distribution networks.
\item Enhanced Training: GMC-based data augmentation improves training diversity and DRL performance.
\item Efficient Solver: Laurent power flow solver reduces computation time tenfold, maintaining accuracy.
\end{highlights}


\maketitle

\section{Introduction}
\subsection{Motivation}
Energy Storage Systems (ESSs) play a pivotal role in modern distribution networks, offering enhanced flexibility essential for addressing uncertainties brought by Distributed Energy Resources (DERs) integration~\cite{specht2023DRLEnergyAssets}. Optimizing ESS dispatch strategies is crucial for distribution system operators (DSOs) to fully harness this flexibility~\cite{pedro2019_voltage_control}. However, the dynamic and sequential nature of optimal operation decisions, responding to fluctuating prices and varying electricity demands, poses a significant challenge. Traditional model-based approaches often struggle with real-time decision-making due to their reliance on predefined forecasts or complex probability functions to manage uncertainties~\cite{shengrenOptimalEnergySystem2023}. 
Deep Reinforcement Learning (DRL) emerges as a potent model-free solution for such fast-paced, sequential decision-making scenarios, with successful applications in diverse fields like game-playing~\cite{gym}, robotics control~\cite{drone-control}, industry control~\cite{2022magnetic}. Applied to distribution energy systems, DRL transforms these operational challenges into a Markov Decision Process (MDP), exhibiting impressive results in various energy tasks~\cite{gallego2023DRL,DRL2024decentralized_adn,pedro2022_rl_votlage_control}. DRL’s strength lies in its adaptability and capability for real-time decision-making, trained in simulators and then applied to real-world scenarios. This necessitates robust and accurate simulation environments to prevent duplication and provide benchmark frameworks for the development of efficient DRL algorithms.

Therefore, we introduce RL-ADN, an open-source library specifically tailored for DRL-based optimal ESSs operation in distribution networks. It meets diverse research needs while providing customization options for research tasks, ensuring both flexibility and standardization.

\subsection{Related Work}

The RL field has grown significantly, thanks in part to open-source universal simulation environments and benchmark frameworks, like GYM for game-playing~\cite{gym}. However, this trend is less pronounced in energy system research groups. The absence of such resources hampers the development and integration of DRL algorithms in energy system operation areas. Table~\ref{summary_literature_review} offers a comparative analysis of functionalities in open-sourced energy system environments. Many existing environments address specific challenges but are often too tailored for broader application~\cite{hou2023constraint,shengren2022performance}. For instance, a microgrid environment is developed to test the performance of DRL algorithms in~\cite{shengren2022performance}. The task of formulated MDP is to minimize the power unbalance and operational cost by dispatching distributed generators and ESSs. In the research~\cite{wang2021multi}, a distribution network environment is open-sourced to facilitate solving active voltage control problems based on multi-agent RL algorithms. AndesGYM~\cite{cui2022andes_gym} developed an environment for frequency control problems in power systems, which leverages the modeling capability of ADNES and Gym environment. The task is set to minimize the deviations of the frequency value in a given time scope. Consequently, these environments do not lend themselves easily to customization or alterations essential for different or broader research objectives. This specificity leads to fragmentation in the research community, as studies operate in isolation without a standardized benchmark or a universally adaptable toolset.

CityLearn~\cite{citylearn} provides an environment for simulating DRL algorithms in charge of operating building energy systems, in either a centralized (single-agent) or decentralized (multi-agent) way. Focusing on exploring the dynamics inside the building, it ignored the grid-level dynamic. GridLearn~\cite{gridlearn} is further developed to investigate mitigating over-voltages in the distribution network level by demand response in the buildings. Both two packages simplified the original MDP tasks, by discrete continuous decisions into discrete actions and ignoring the power flow calculation in the distribution networks. PowerGridWorld~\cite{biagioni2022powergridworld} is a framework for researchers to customize multi-agent environments of power networks, which could integrate existing RL libraries like RLLIb and OPEN-AI BASELINES. PowerGridWorld could work in two ways to implement the multi-agent feature: centralized training and distributional execution, distributional training, and execution. In the environment, OPENDSS is used as an interface to execute the power network operation. Gird2OP~\cite{grid2op} is developed to support training an intelligent agent to run a transmission network and has served as a benchmark environment for a series of L2RPN competitions. Grid2OP provided the flexibility for grid modifications, observations, and actions. However, both PowerGridWorld and Grid2OP necessitate extensive power flow calculations during offline training, typically a bottleneck in DRL training, since RL agents need to explore the environments to converge, requiring a large amount of interaction. The mentioned electricity network environments are mainly built based on standard iterative methods, i.e. Newton-Raphson method, which is time-consuming, rendering them unsuitable for integration with DRL algorithms training. GYM-ANM~\cite{gym-anm} is an open-source environment for solving operation problems in distribution networks, with the primary purpose of using RL algorithms to reduce energy loss (including generation curtailment storage, and transmission losses) under the operation violation constraints. GYM-ANM provides flexibility for customizing energy components, research tasks, network topology, etc. Specifically, it uses a customized simplified power flow simulator to encapsulate the dynamics of a distributional network, which can accelerate the training speed of RL agents significantly. However, the limitations of GYM-ANM are also obvious, as the implemented power flow calculation algorithm can not precisely track the dynamic of physical distribution networks, impeding the transition from simulation to reality for the trained RL agents. Therefore, an advanced power flow calculation algorithm remains a significant imperative to avoid being hindered by the extensive computational demands as well as to reflect the dynamics of physical distribution networks accurately.

Moreover, the key to leveraging DRL for optimal dispatch strategies lies in training with diverse historical data, particularly in environments with uncertain renewable generation, load consumption, and price profiles. The broader the training scenarios, the higher the DRL agents' performance ceiling~\cite{shengren2022performance}. However, collecting diverse data for specific distribution networks remains challenging, limiting the practical integration of DRL algorithms.

\begin{table*}
\centering
\caption{Summary of literature in environments of distribution network operation. The content of the table strictly aligns with the novelty we include: power flow integration, data augmentation, benchmark optimality, and flexibility assessment.}
\label{summary_literature_review}
\scalebox{0.75}{
\begin{tabular}{lllll} 
\hline
\textbf{Work} & \textbf{Research Task} & \textbf{Power Flow Integration} & \textbf{Data Augmentation} & \textbf{Flexibility and Customization Capabilities} \\ 
\hline
\cite{shengren2022performance} & Optimal energy system scheduling & \(\times\) & \(\times\) & \(\times\) \\
\cite{wang2021multi} & Voltage regulation & \checkmark & \(\times\) & \(\times\) \\
CityLearn~\cite{citylearn} & Building Energy Management & \(\times\) & \(\times\) & \checkmark\\
GridLearn~\cite{gridlearn} & Building Energy Management & \(\times\) & \(\times\) & \checkmark\\
PowerGridWorld~\cite{biagioni2022powergridworld} & Power Network Operation & \checkmark & \(\times\) & \checkmark\\
Grid2OP~\cite{grid2op} & Transmission Network Configuration & \checkmark & \(\times\) & \checkmark\\
GYM-ANM~\cite{gym-anm} & Distribution Network Operation & \checkmark & \(\times\) & \checkmark\\
\cite{shengren2023optimal} & Microgrid operation & \(\times\) & \(\times\) & \(\times\) \\
\cite{xu2022EV_DRL} & EV energy management & \(\times\) & \(\times\) & \(\times\) \\
\cite{bode2020scalable} & Microgrid Control & \checkmark & \(\times\) & \(\times\) \\
\cite{lerousseau2021design} & Microgrid operation & \checkmark & \(\times\) & \checkmark \\
\cite{de2021applying} & Economic dispatch & \(\times\) & \(\times\) & \(\times\) \\
\cite{huang2019adaptive} & Power system emergency control & \checkmark & \(\times\) & \checkmark \\
\cite{cui2021decentralized} & Voltage Control & \checkmark & \(\times\) & \(\times\) \\
\midrule
\textbf{RL-ADN} & Optimal ESSs dispatch in distribution network & \checkmark & \checkmark & \checkmark \\
\hline
\end{tabular}}
\end{table*}

\subsection{Contributions}

This paper presents RL-ADN, an open-source library for DRL-based optimal ESSs dispatch in active distribution networks. RL-ADN accommodates a wide range of research objectives (i.e., different optimization objectives functions such as congestion management and optimal dispatch) while offering unprecedented customization capabilities. This flexibility extends to the modeling of distribution network topologies and the integration of various types of ESSs, thereby allowing for the creation of tailored MDPs. RL-ADN incorporates a novel data augmentation module using a Gaussian Mixture Models-Copula (GMC) approach, enhancing the diversity of training scenarios and thereby the performance of DRL algorithms. Additionally, it introduces the Laurent power flow solver, drastically reducing computation time for power flow calculations tenfold, without sacrificing accuracy~\cite{juan-tensorpower,duque2024tensor}. RL-ADN also provides four state-of-the-art (SOTA) DRL algorithms and a model-based approach with perfect forecasts as a standard baseline for comparison. In summary, RL-ADN sets a new standard in DRL-based ESS dispatch with its innovative features, flexibility, and efficiency. It paves the way for more effective and accurate DRL applications in energy distribution networks, representing a significant advancement in the field.

\section{Background}
\subsection{Optimal ESS dispatch tasks in distribution networks}

ESSs dispatch tasks are inherently sequential decision-making problems. The aim is to minimize operational costs while adhering to constraints that ensure the safe and efficient operation of the distribution network. Such constraints might include maintaining specific voltage magnitude and current levels, state of charge (SOC) operation constraints, etc. This involves responding to market prices, network conditions, and renewable stochastic generation. The ESSs dispatch problem is typically cast as optimization problems with a general mathematical optimization formulation defined by \eqref{eq:goal}--\eqref{eq:ess_const}:

Minimize:
\begin{equation}\label{eq:goal}
    f(x) \quad \text{where } x \text{ is the decision variable.}
\end{equation}

Subject to:
\begin{flalign}
& g(x) < y & \quad \text{(Grid-level constraints)} \label{eq:grid_const}\\
& b(x) < z & \quad \text{(Energy storage system constraints)} \label{eq:ess_const}
\end{flalign}

The objective function \( f(x) \) varies based on different tasks, ranging from minimizing operation cost based on dynamic pricing to regulating voltage magnitude or integrating multiple goals~\cite{mauricio2022eligibility}. The effective dispatch of ESSs is crucial, considering the uncertainties in renewable generation, load consumption, and price fluctuations. The constraints are categorized into grid-level \eqref{eq:grid_const} and ESS-level \eqref{eq:ess_const} based on the specific requirements of the tasks. Some tasks may prioritize network reliability and incorporate more stringent constraints on voltage magnitude and current levels, while others may focus solely on profit maximization. This flexibility in formulation allows for a wide array of approaches, each tailored to meet the specific needs and priorities of different energy optimization tasks.

\subsection{MDP formulation and reinforcement learning}

In RL-ADN, these sequential decision-making problems can be reformulated as a MDP, defined by the tuple \( (\mathcal{S},\mathcal{A},\mathcal{P},\mathcal{R},\gamma) \), where \(\mathcal{S}\) denotes the state space, \(\mathcal{A}\) represents the action space, \(\mathcal{P}\) is the state transition probability function, \(\mathcal{R}\) signifies the reward function, and \(\gamma\) stands for the discount factor.

A policy, \( \pi(a_t|s_t) \), determines the selection of action \( a_t \) for a given state \( s_t \). The agent's objective is to ascertain a policy that maximizes the expected discounted cumulative return, represented as $J(\pi)=\mathbb{E}_{\tau \sim \pi}\left[\sum_{t=0}^{{\cal T}} \gamma^t r_t\right]$, in which ${\cal T}$ is the length of the control horizon.

\begin{figure*}
    \centering
    \psfrag{M1}[][][0.8]{$\pi_{\omega}(s)$}
    \psfrag{M2}[][][0.8]{$Q_{\theta}(s,a)$}
    \psfrag{M3}[][][0.8]{$\pi_{\omega}(s)$}
    \psfrag{M4}[][][0.8]{$Q_{{\theta}_1}(s,a)$}
    \psfrag{M5}[][][0.8]{$Q_{{\theta}_2}(s,a)$}
    \psfrag{M6}[][][0.8]{$\pi_{\omega}(s)$}
    \psfrag{M7}[][][0.8]{$V_{\phi}(s)$}
    \psfrag{M8}[][][0.8]{$\pi_{\omega}(s)$}
    \psfrag{M9}[][][0.8]{$Q_{{\theta}_1}(s,a)$}
    \psfrag{M10}[][][0.8]{$Q_{{\theta}_2}(s,a)$}
    \psfrag{K1}[][][0.8]{$V_{\phi}(s)$}

    \includegraphics[width=1.0\linewidth]{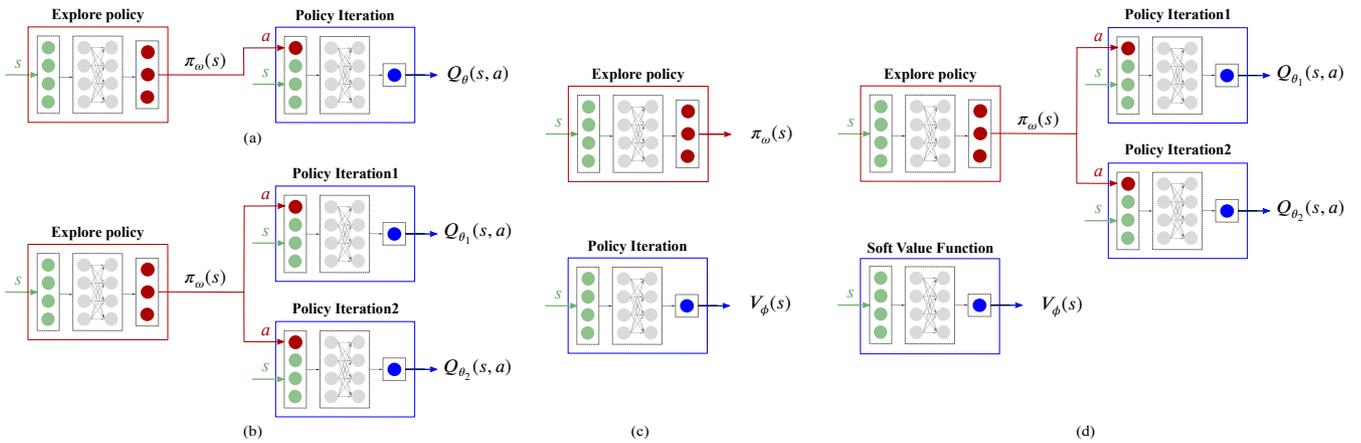}
    \caption{Architecture of policy-based DRL algorithms. (a) Deep Deterministic Policy Gradient (DDPG), (b) Twin Delayed DDPG (TD3), (c) Proximal Policy Optimization (PPO), (d) Soft Actor-Critic (SAC).}
    \label{fig:DRLs}
\end{figure*}

The formulated MDP possesses a continuous action space, making it unsuitable for direct solutions using value-based DRL algorithms~\cite{sutton_reinforcement_2018}. Policy-based DRL algorithms are often employed to address continuous action spaces, as they directly tackle such continuous action domain problems. The architectures of state-of-the-art (SOTA) policy-based DRL algorithms such as DDPG~\cite{DDPG}, TD3~\cite{fujimoto_td3_2018}, SAC~\cite{haarnoja_sac_2018}, and PPO~\cite{schulman_ppo_2017} are depicted in Fig.~\ref{fig:DRLs}.

\begin{itemize}
    \item \textbf{DDPG and TD3:} Both are deterministic algorithms that maintain a policy for action sampling and Q-networks, \( Q_\theta(s_t,a_t) \), to guide policy network updates. Specifically, TD3, as an enhancement of DDPG, incorporates dual Q-networks and employs delayed updates, mitigating the Q-network's overestimation bias inherent in DDPG.
    
    \item \textbf{PPO:} As an on-policy algorithm, PPO addresses policy optimization challenges in RL. PPO curtails extensive policy updates by adopting a clipped objective function, ensuring minimal deviation of the new policy from the previous one. A value function $V_{\phi}(s)$ is leveraged to guide the policy iteration. This mechanism circumvents the necessity of learning rate adjustments and achieves superior sample efficiency compared to conventional policy gradient techniques~\cite{schulman_ppo_2017}.
    
    \item \textbf{SAC:} SAC is an off-policy actor-critic framework that integrates the maximum entropy reinforcement learning paradigm. By supplementing the typical reward with an entropy component, SAC promotes exploration, thereby achieving a harmonious balance between exploration and exploitation. This algorithm utilizes a soft value function, dual Q-functions, and a policy network. With iterative updates, SAC strives to formulate a stochastic policy that is both optimal and exploratory, ensuring robustness and efficiency across diverse tasks.
\end{itemize}

Building on the policy gradient theorem, both the policy, \( \pi(a_t|s_t) \), and its associated critic networks, \( Q_\theta(s_t,a_t) \) or $V_{\phi}(s)$, can be updated. It is worth noting that the update methods can vary depending on the specific algorithm. A comprehensive discussion of these algorithms is available in~\cite{chen_reinforcement_2021}. 


By interacting with the artificial environment, the DRL agent seeks to define the optimal ESSs dispatch in active distribution networks. The two-phase approach, offline training followed by online deployment, equips the agent to address the stochastic nature in optimal ESSs dispatch tasks. In the offline training phase, the DRL agent gleans insights from the interaction and executes self-learning, refining its decision-making. During the subsequent online deployment, it leverages these insights to navigate complexities, ensuring more robust and adaptive solutions. The environment's partially observable nature, often due to communication constraints, necessitates meticulous state selection from the full observation set. Overly complex states will decrease the signal-to-noise ratio, while overly simplistic states could overlook essential dynamics. Both scenarios can undermine the learning efficacy and policy performance. To provide flexibility in designing state spaces, RL-ADN facilitates the easy customization of state spaces, a topic further explored in the subsequent sections.

\section{RL-ADN Framework}
\subsection{Overview}

\begin{figure}
    \centering
    \psfrag{A1}[][][0.8]{$s_t$}
    \psfrag{A2}[][][0.8]{$r_{t-1}$}
    \psfrag{A3}[][][0.8]{$a_t$}
    \psfrag{A4}[][][0.8]{$s_{t+1}$}
    \psfrag{A5}[][][0.8]{$s_{t+1}$}
    \psfrag{A6}[][][0.8]{$r_{t}$}
    \psfrag{A7}[][][0.8]{$[s_{t+1},r_{t}]$}
    \includegraphics[width=1.0\columnwidth]{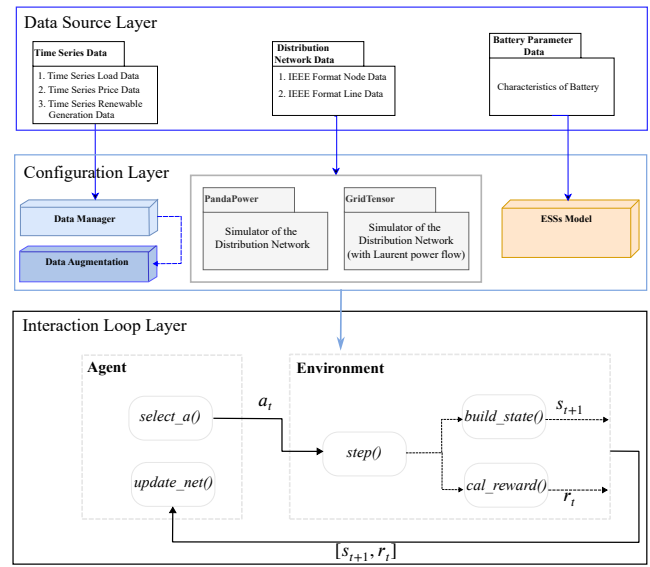}
    \caption{Framework of the RL-ADN package. Configuration data for the distribution network and the ESSs are selected from data sources. Subsequently, corresponding time-series data undergo preprocessing. Through Configuration Layer, the environment is constituted of the distribution network, ESSs, and data manager.}
    \label{fig:RL-ADN framework}
\end{figure}

The architecture of the RL-ADN environment, depicted in Fig.~\ref{fig:RL-ADN framework}, consists of three layers: Data Source, Configuration, and Interaction Loop. Primary data feed into Configuration Layer to build DRL environments, integrating components like Data Manager, Distribution Network Simulator, and ESSs Models. These components are integrated into the environment within the Interaction Loop, while a DRL algorithm, chosen to control the agent, is initialized simultaneously\footnote{State-of-the-art policy-based algorithms such as DDPG, SAC, TD3, and PPO are incorporated into the framework.}. Then, the DRL agent interacts with the environment in search of the optimal policy. The proposed RL-ADN framework's versatility allows for modeling highly tailored tasks, with modifications to components yielding unique MDPs for distinct ESSs dispatch tasks.

\subsection{Data Source Layer}

The Data Source Layer provides primary data for building the framework and training the DRL agent. Data are categorized into time-series data, distribution network data, and ESSs parameter data. Time-series data include load profiles, price profiles, and renewable generation profiles in a standard format. These data are processed by the Data Manager for training or can be selected for further augmentation. Distribution network data comprise node and line data, with nodes specifying slack and PQ bus locations, and lines detailing topology and characteristics like resistance and reactance which are stored in CSV format. This data is crucial for building the distribution network simulator. ESSs parameter data, detailing capacity, charge/discharge limits, and degeneration costs, are used to construct the ESSs model. The framework includes standard 25, 34, 69, and 123 node distribution network data, along with corresponding time-series data and ESSs data from previous research\cite{hou2023constraint}. Users can use this data for training or customize their own model following the provided standard format. 

\subsection{Configuration Layer}
\subsubsection{Data Manager}
The Data Manager plays a crucial role in managing time-series data, such as active and reactive power demand (\( p^D_{i,t} \), \( q^D_{i,t} \)), electricity price (\( \rho_t \)), and renewable power generation (\( p^R_{i,t} \), \( q^R_{i,t} \)) for specific epochs (\( \mathcal{T}, t\in \mathcal{T} \)). Previous research approaches to data management have been case-specific and labor-intensive, adding complexity and potential data quality issues. RL-ADN adopts a streamlined approach, standardizing various data preprocessing tasks, and ensuring data integrity and efficient handling. The workflow of the Data Manager is detailed in Appendix~\ref{sec_data_manager_workflow}.

\subsubsection{Data Augmentation}
In RL-ADN, Data Augmentation module plays a pivotal role in enhancing the robustness and generalizability of the trained policy by artificially expanding the diversity of the historical time-series data.
With data augmentation, RL-ADN exposes the model to a broader set of scenarios, promoting adaptability and performance in varied and unforeseen situations.
The Data Augmentation module is designed to generate synthetic time-series data, capturing the stochastic nature of load in the power system and reflecting realistic operational conditions. The Data Augmentation module interacts with the Data Manager to retrieve the necessary preprocessed data and then applies its augmentation algorithms to produce an augmented dataset. The output is a synthetic yet realistic dataset that reflects the variability and unpredictability inherent in distribution network systems. This enriched dataset is crucial for training RL agents, providing them with a diverse range of scenarios to learn from and ultimately resulting in a more adaptable and robust decision-making policy. The workflow of Data Augmentation module is described in Appendix~\ref{sec_data_augmentation_workflow}.

\subsubsection{Distribution Network Simulator}
For a distribution network, node-set \( \mathcal{N} \) and the line set \( \mathcal{L} \) define the topology. Each of the node \( i\in \mathcal{N} \) and lines \( l_{i,j}\in \mathcal{L} \) specify its attributes. A specific subset \( \mathcal{B},\mathcal{B}\subset \mathcal{N} \) describes ESSs connected to the distribution network nodes. Importantly, the number of ESSs delineates the resulting state space \( \mathcal{S} \) and action space \( \mathcal{A} \).

The main function of the Distribution Network Simulator is to calculate power flow, when a new scenario is fed into the environment, performing as the main part of the state transition function for the formulated MDP task. Based on the provided distribution network configuration data, we offer two modules, \texttt{PandaPower} and \texttt{GridTensor} to create the Distribution Network Simulator. \texttt{PandaPower} provides the traditional iterative methods while \texttt{GridTensor}~\cite{juan-tensorpower} integrates a fast Laurent power flow for calculating the distribution network state presented by the voltage magnitudes, currents and power flowing in the lines. 

\subsection{Interaction Loop Layer}

For each time step $t$ in an episode, the agent obtains the current state $s_t$ and determines an action $a_t$ to be executed in the environment. Once $a_t$ is received, the environment will execute \texttt{step} function to execute power flow, and update the status of ESSs and the distribution network, which is counted as the consequence of the action at the current time step $t$. Then, based on these resultant observations, the reward $r_t$ is calculated by the designed reward calculation block. Next, the Data Manager in the environment samples external time-series data of the next time step $t+1$, including demand, renewable energy generation, and price, emulating the stochastic fluctuations of the environment. These external variables are combined with updated internal observations, performing as the resultant transition of the environment. 

Users can freely design the \texttt{build-state} block, facilitating an in-depth exploration of how different states influence the performance of algorithms on various tasks. In a similar vein, the \texttt{cal-reward} block can be tailored according to different optimal tasks. For the convenience of our users, our framework provides a default state pattern and reward calculation.

\subsection{MDP Design}

\subsubsection{State Space Design}

State space design is vital as it directly impacts the efficacy of the agent's learning process. The chosen state space \( \mathcal{S} \) should be concise yet descriptive enough to facilitate effective policy learning.

In the RL-ADN framework, the environment collect a comprehensive range of measurements at each timestep \( t \). Using all these measurements to represent the state \( s_t \) in the MDP is plausible but fraught with challenges. Such an approach might not be practical in real-world distribution networks due to potential data unavailability. Moreover, by including all measurements, the state space could become noise-prone, making state exploration more intricate and possibly hindering agent performance.

Thus, feature engineering is pivotal in designing state $s_t$. The RL-ADN framework offers the flexibility to tailor state space. The \texttt{get-obs} block fetches available measurements, while the \texttt{build-state} block lets users customize states. Generally, the state $s_t$ encompasses both endogenous and exogenous features. Exogenous features capture external dynamics, like uncertainties in renewable energies, consumption, and pricing, within an episode. Meanwhile, endogenous features track internal dynamics governed by distribution network rules and energy component behaviors, e.g., power flow and ESS's SOC update rules. Moreover, some ancillary information, such as the current time-step in a trajectory, has proven crucial in MDP state representation~\cite{mauricio2022eligibility}.

\subsubsection{Action Space Design}

Focusing the optimal ESS dispatch tasks, the action \( a_t \) at time \( t \) is denoted as \( a_t = [p^{B}_{m,t}|_{m \in \mathcal{B}}] \), symbolizing the charging or discharging directives for the \( m_{th} \) ESS connected to node \( m \) in the distribution network.

\subsubsection{Transition Function}

In a MDP, the transition function encapsulates the dynamics that govern the system's progression from one state to another. The transition mechanism is bifurcated into two essential components. The first is endogenous distribution network and energy component dynamics. These are calculated based on physical laws, i.e. power flow calculation, SOC update rules, rooted in the network's topology, the variations in active and reactive power at different nodes, and the parameter of ESSs models. The second is exogenous variable evolution, which involves modeling the temporal fluctuations in renewable energy generation, market prices, and load demand, leveraging daily historical data. The transition probability function \( \mathcal{P} \) is mathematically represented as:

\begin{multline}
    \label{eq:transition_function}
  p(S_{t+1},R_{t}|S_t,A_t) = \\\operatorname{Pr}\left\{S_{t+1}=s_{t+1}, R_{t}=r_{t} \mid S_{t}=s_{t}, A_{t}=a_t\right\}.
\end{multline}

Traditionally, constructing a precise mathematical representation of \( \mathcal{P} \) has been challenging due to the inherent complexities and uncertainties in both endogenous and exogenous variables. Reinforcement Learning (RL) offers a way around this by learning the ambiguous model through interaction\footnote{Model-free RL algorithms obviate the need for explicit knowledge of \( \mathcal{P} \), enabling the agent to learn optimal policies through interaction with the environment.}.


\subsubsection{Reward Function}
\label{sec_reward_design}

The reward function serves as a critical component for guiding the agent's learning process. The environment offers a reward signal \( r_{t} \) to the agent, quantifying the quality of each action taken. The design of this reward function is inherently tied to the specific objectives of the task at hand~\footnote{The default reward functions are presented in Section 4.1.}. Our framework incorporates a \texttt{cal-reward} block that allows researchers to easily customize the reward signal for various optimal ESS dispatch challenges.

\subsection{Data Augmentation Model}

The RL-ADN framework incorporates Gaussian mixture models (GMM) and Copula functions for data augmentation \cite{bernards2017statistical, duque2021conditional}. The GMM is a probabilistic model that assumes data originates from a blend of multiple Gaussian distributions, each characterized by unique means and covariances. This model can adeptly capture the complex and multi-modal nature of time series data in distribution networks, which often exhibit intricate patterns due to fluctuating load demands and renewable energy generation. Complementing the GMM, Copula functions are utilized to encapsulate the time-correlation structure between multiple time-step data in a defined period, independent of their marginal distributions. This dual approach ensures a comprehensive and realistic augmentation of time-series data in distribution network operations. In our framework, three augmentation methods are provided GMM, t-Copula, and Gaussian Copula~\cite{xia2024comparative}.

The integration of GMM and Copula functions (GMC) in the RL-ADN framework marks a significant advancement in creating robust and reliable environments for training reinforcement learning agents. This approach adeptly handles the complexities and uncertainties inherent in power distribution networks, enhancing the training data's quality and the resulting policies' effectiveness.

\subsection{Laurent Power Flow}
\label{sec:fast_power_flow}

Conventional power flow calculations often rely on iterative methods like the Newton-Raphson algorithm. This becomes a computational bottleneck, especially in the context of training DRL agents, which requires numerous evaluations of power flow. In the proposed framework, we address the computational bottleneck associated with traditional power flow calculations, by incorporating a Laurent power flow algorithm~\cite{juan-tensorpower}. This efficiency approach is achieved by linearizing the power flow equations using a Laurent series expansion, which simplifies the nodal current calculations in the distribution network. By doing so, we facilitate frequent power flow evaluations necessary for training RL agents, without the computational burden.

The Laurent power flow method we employ considers both constant power and constant impedance loads, integrating the ZIP load model directly into the power flow analysis. This approach allows for the inclusion of various types of loads and renewable energy sources without the need for iterative approximation methods typically used in traditional power flow analysis. As a result, our algorithm achieves rapid convergence and permits a more streamlined and scalable RL training process. The elimination of iterative computation not only expedites the power flow assessment but also enhances the RL agent's ability to quickly adapt and learn, thereby improving the overall efficiency and effectiveness of the framework.

\section{Benchmark Scheme and Experiments}
\subsection{Optimal ESSs dispatch Task and MDPs}

\begin{figure}[t]
\centering
  \psfrag{A1}[lc][][0.8][0]{Load bus}
  \psfrag{A2}[lc][][0.8][0]{PV generation}
  \psfrag{A3}[lc][][0.8][0]{ESS}
\includegraphics[width=1.0\columnwidth]{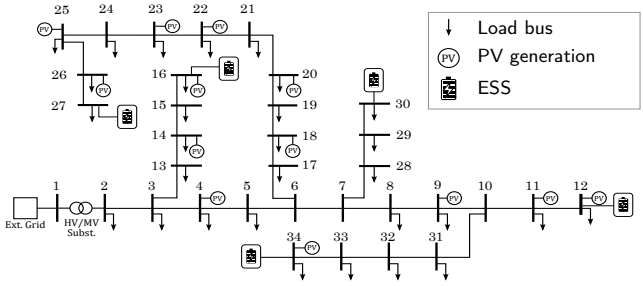}
\vspace{-2mm}
\caption{Modified IEEE-34 Node bus test system with distributed PV generation and EESs. The ESSs are placed at the end of each feeder to increase the number of voltage magnitude issues experienced.}
\vspace{-2mm}
\label{fig:ieee_system}
\end{figure}

RL-ADN framework introduces a foundational optimal ESSs dispatch case while the mathematical formulation of the case is shown in Appendix~\ref{sec_math_formulation}. This default case aims to minimize the operational costs for DSOs while ensuring compliance with the distribution network and ESSs operation constraints. The template case offers researchers and practitioners a springboard, enabling them to design bespoke benchmarks tailored to unique ESSs dispatch challenges.

In the provided case, a modified 34-node IEEE test distribution network is leveraged to build the Distribution Network Simulator, as illustrated in Fig.~\ref{fig:ieee_system}. Strategic placement of the ESSs on nodes 12, 16, 27, and 34 which have over- and under-voltage issues. The objective remains to minimize the operational cost, while upholding voltage magnitude constraints. Consequently, the state, and reward functions are constructed as below: 
the state \( s_t \) is described as \( s_t = [P^{N}_{m,t}|_{m \in \mathcal{N}}, \rho_t, SOC^{B}_{m,t}|_{m \in \mathcal{B}}] \), incorporating both endogenous and exogenous features. The design of \( \mathcal{A} \) adheres to the optimal goal and multiple constraints:

\begin{itemize}
    \item \textbf{Charge and Discharge Bounds:} ESSs have inherent physical limitations. The action \( a_t \) is confined within a range, considering these physical constraints.
    \item \textbf{State-of-Charge (SOC) Dependency:} Actions must respect the current SOC of each ESS. The `step' function ensures this by adjusting the charge/discharge commands based on SOC levels.
    \item \textbf{Voltage Magnitude Regulation:} ESS actions should maintain voltage within predefined limits. Direct enforcement is infeasible; hence, we employ soft constraints via penalty rewards for voltage violations.
\end{itemize}

Thus, the reward function is defined as the combination of energy arbitrage profits and the penalty of the voltage magnitude violations in the distribution network. Mathematically, this is expressed as:

\begin{multline}\label{eq:new_penalty_reward}
r_t =  \rho_{t} \left[ \sum_{m \in {\cal N}} \left(P^{B}_{m,t} \right) \right] \Delta t  
-\sigma \left[ \sum_{m \in {\cal B}} C_{m,t}(V_{m,t}) \right],
\end{multline}
where \( C_{m,t} \) is constraint violation functions~\cite{shengrenOptimalEnergySystem2023}: \begin{equation}
    C_{m,t}=\min \left\{0, \left(\frac{\overline{V}-\underline{V}}{2} - \left|V_0 - V_{m, t}\right|\right)\right\}, \forall m \in {\cal B}.
\end{equation}
where \( \sigma \) is a trade-off parameter between energy arbitrage and voltage stability.

\subsection{Bench-marking Approach}

To assess performance, we formulate the optimal ESS dispatch problem as a model-based optimization problem, with ESS dispatch decisions as the primary variables. Historical data — including renewable generation, load consumption, and market prices — are treated as perfect forecasts and inputted into the optimization model. Solving this model yields a globally optimal solution, serving as a benchmark for evaluating DRL-derived strategies. Following previous research~\cite{hou2023constraint}, we can assess the efficiency of DRL algorithms by defining performance bound:
\begin{equation}
\text{Performance Bound} = \frac{C_{DRL} - C_{opt}}{C_{opt}}
\end{equation}
Where $C_{DRL}$ is the operational cost of the dispatch strategy derived from DRL agents, while $C_{opt}$ is that derived from the global optimal solution. The closer the DRL decisions align with this benchmark, the higher the efficacy of the RL agents. We incorporate SOTA DRL algorithms capable of handling continuous action spaces, such as DDPG, PPO, SAC, and TD3, as our benchmark DRL algorithms.

Following prior research \cite{shengrenOptimalEnergySystem2023}, our simulation dataset comprises electricity market prices from the Netherlands, augmented with consumption and PV generation data at a 15-minute resolution. Hyperparameter settings for the utilized DRL algorithms are detailed in Table~\ref{table:hyperparamters}. We compare the performance of these DRL algorithms against global optimal solutions obtained by formulating Nonlinear Programming (NLP) problems, solved using the Pyomo package \cite{hart_pyomo_2017}.

\begin{table}[]
\caption{Summary - Parameters for DRL algorithms and the MDP}\label{table:hyperparamters}
\centering
\scalebox{0.95}{
\begin{tabular}{cc}
\toprule
\multirow{4}{*}{PPO   Alg.} & $\gamma=0.995$\\
                           & $\text{Optimizer = Adam}$\\
                           & $\text{Learning rate=}6e-4$\\
                           & $\text{Batch size}=4096$ \\
                           &
                           $\text{GAE parameter} (\lambda)=0.99$ \\
                           \hline
\multirow{4}{*}{DDPG, TD3   Alg.} & $\gamma=0.995$\\
                           & $\text{Optimizer = Adam}$\\
                           & $\text{Learning rate=}6e-4$\\
                           & $\text{Batch size}=512$ \\
                           &
                           $\text{Replay buffer size} =4e5$ \\
                           \hline
\multirow{4}{*}{SAC   Alg.} & $\gamma=0.995$\\
                           & $\text{Optimizer = Adam}$\\
                           & $\text{Learning rate=}6e-4$\\
                           & $\text{Batch size}=512$ \\
                           &
                           $\text{Entropy=auto}$ \\
                           \hline
Reward                     & $\sigma=400$ \\ \hline

\multirow{2}{*}{ESSs}& $\overline{p}^B=50kW,\underline{p}^B=-50kW$, \\
&$\overline{SOC}^B=0.8,\underline{SOC}^B=0.2$, \\ \hline
Voltage limit & $\overline{v}=1.05,\, \underline{v}=0.95$\\
\bottomrule
\vspace{-6mm}
\end{tabular}}\label{tab:table_summ}
\end{table}

\section{Results}
\subsection{Performance of DRL Algorithms on Template Optimal Dispatch Task}

\begin{figure}[t]
    \psfrag{Total Reward [-]}[][][1.0]{$\text{Total Reward [-]}$}
    \psfrag{A3}[][][0.2]{$\sum_{m \in {\cal N}} \left( P^D_{m,t}+P^{B}_{m,t}-P^{PV}_{m,t}\right)\text{[-]}$}
    \psfrag{A4}[][][0.2]{$\sum_{m\in\cal{B}} C_{m,t}(V_{m,t})\text{[-]}$}
    \centering
\includegraphics[width=1.0\columnwidth]{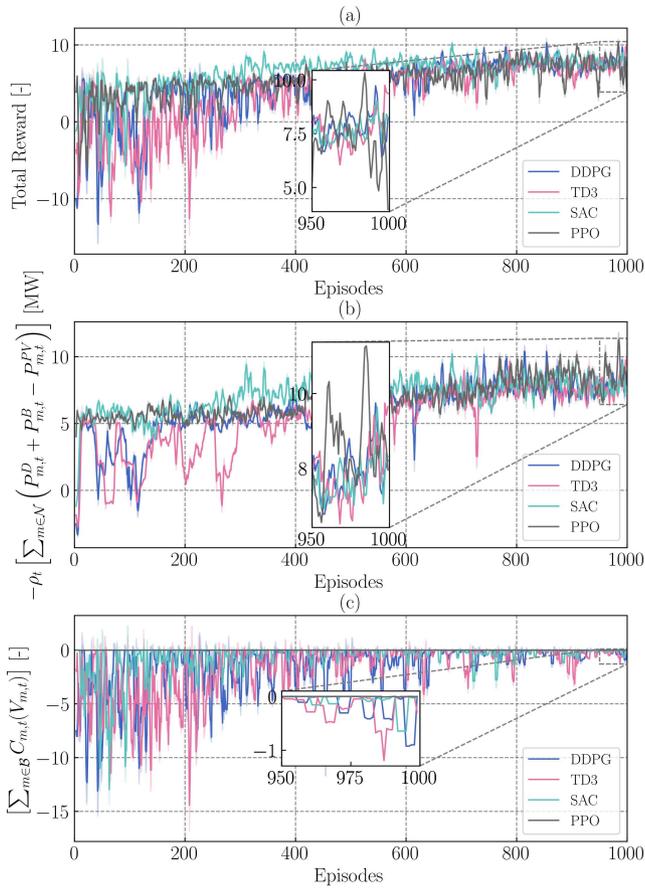}
    \caption{$(a)$ Average total reward as in \eqref{eq:new_penalty_reward}. $(b)$ Operational cost or first term of reward in \eqref{eq:new_penalty_reward}. $(c)$ Cumulative penalty for voltage magnitude violations or second term of reward in \eqref{eq:new_penalty_reward}, all during training.}
    \label{fig:training_data}
    \vspace{-3mm}
\end{figure}

\begin{figure}[ht!]
    \centering
    \includegraphics[width=1.0\columnwidth]{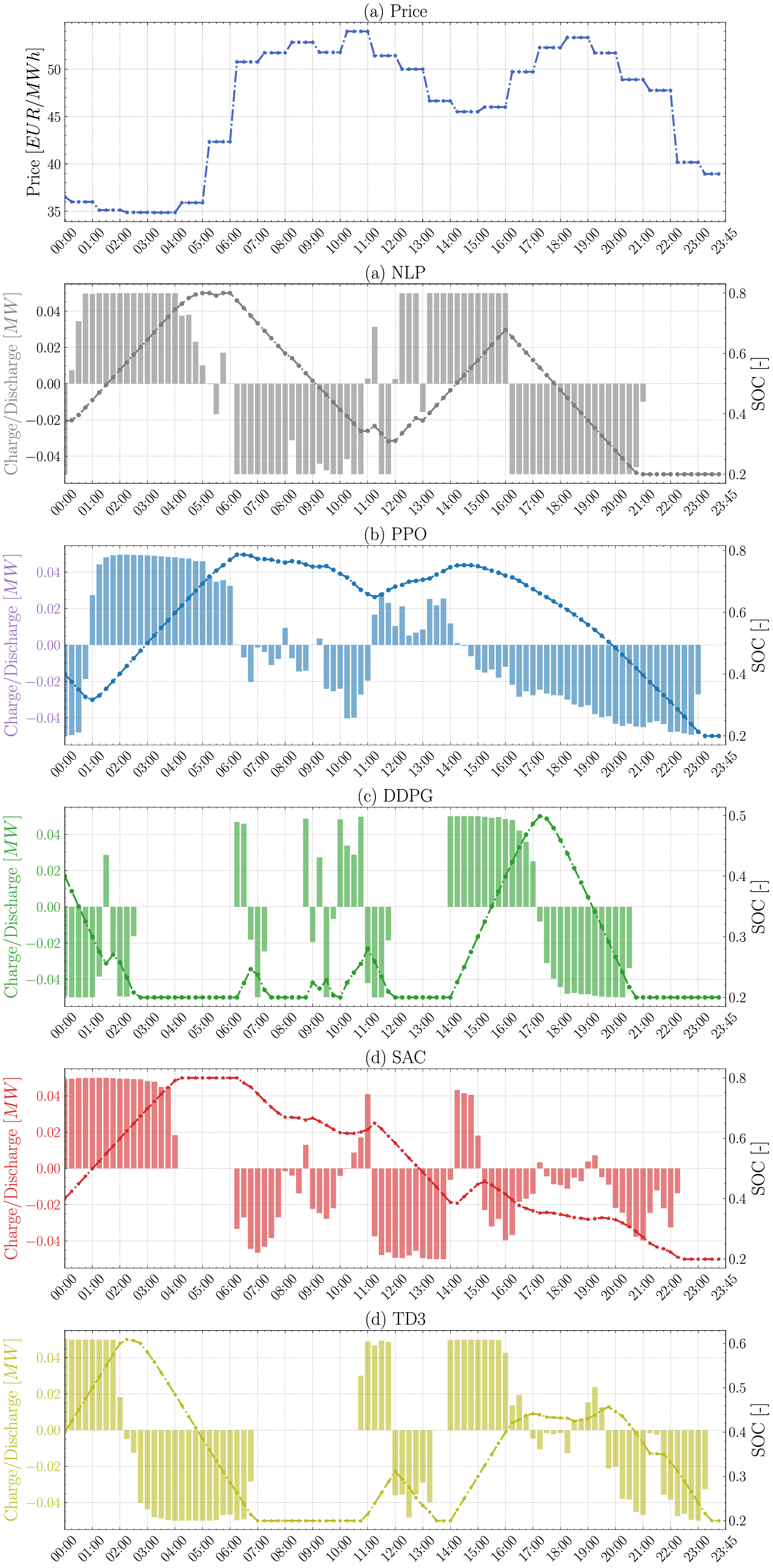}
    \caption{Dispatch decisions obtained by DRL algorithms and NLP for the ESS connected to node 16}
    \label{fig_compare_benchmark}
\end{figure}

Fig.~\ref{fig:training_data} displays the average total reward, operational cost, and the number of voltage magnitude violations during the training process for DDPG, SAC, TD3, and PPO algorithms. Results shown in Fig.~\ref{fig:training_data} are obtained as an average of over five random seeds. The average total reward increases rapidly during the training, while simultaneously, the number of voltage magnitude violations decreases. This is a typical training trajectory of DRL algorithms solving optimal dispatch formulated MDP tasks, especially for those using penalty as a reward. At the beginning of the training process, the DNN's parameters are randomly initialized, and as a consequence, the actions defined usually are random discharge/charge decisions, causing a high number of voltage magnitude violations, thus introducing a huge magnitude penalty term in reward~\eqref{eq:new_penalty_reward}. Such a reward acts as an indicator to guide updating the DNN's parameters, resulting in higher quality actions, primarily learning to reduce voltage magnitude violations. Then, after reducing the violations, DRL algorithms learn to improve the actions toward increasing and minimizing the operational costs. All these DRL algorithms converged at around 1000 episodes. The total reward of these algorithms converged at $7.5\mypm0.02$. Notice that even converged, the operational cost shown in Fig.~\ref{fig:training_data}$(b)$ will not remain the same because the different daily load and price profiles are sampled during the training. After the last training episode, the penalty voltage magnitude violation penalty for these DRL algorithms was reduced to a value of no more than 1 as is shown in Fig.~\ref{fig:training_data}$(c)$. This result shows that DRL algorithms can effectively learn from interactions, reducing the number of voltage magnitude violations while minimizing the operational costs by learning to dispatch the ESSs correctly.

Fig.~\ref{fig_compare_benchmark} shows the dispatch decisions and SOC changes of the ESS, connected to node 16 in a typical daily operation. These decisions are defined by DDPG, TD3, PPO, and SAC, as well as the global optimality benchmark solution provided by solving the NLP formulation considering the perfect forecast. Decisions provided by all DRL algorithms all responded to the dynamic prices during the day. On this day, PPO and SAC perform better than DDPG and TD3. Between 1:00-5:00, when the electricity price is low, PPO and SAC dispatch the ESS in charging mode, which is similar to the decisions from the NLP solver. However, DDPG and TD3 fail to learn to act efficiently with the low prices in these timeslots. During the afternoon, all DRL algorithms charge ESSs between low-price slots while discharging between high-price time slots (see Fig.~\ref{fig_compare_benchmark}$(b)$ and $(c)$). However, Both DRL algorithms fail to capture the price fluctuations perfectly, compared to the decisions from NLP with full observation of the future. For instance, DDPG performs best among all DRL algorithms between 14:00 and 20:00 but fails to capture the price fluctuations well in the morning. PPO generally performs well during the whole day's operation but defines conservative decisions from 6:00 to 14:00. 

Compared to the solution provided by NLP, all DRL algorithms converge to a local optimum after training in the current historical dataset. This performance can be caused by the limited scenarios in the training dataset, which hinder the implication of DRL algorithms in the realistic optimal dispatch operation. In the next section, we show how the performance of DRL algorithms is significantly influenced by using the data augmentation model incorporated in the RL-ADN framework.

\subsection{Impacts of Data Augmentation on Performance of DRL algorithms}

\begin{figure}[ht]
    \centering
    \includegraphics[width=1.0\columnwidth]{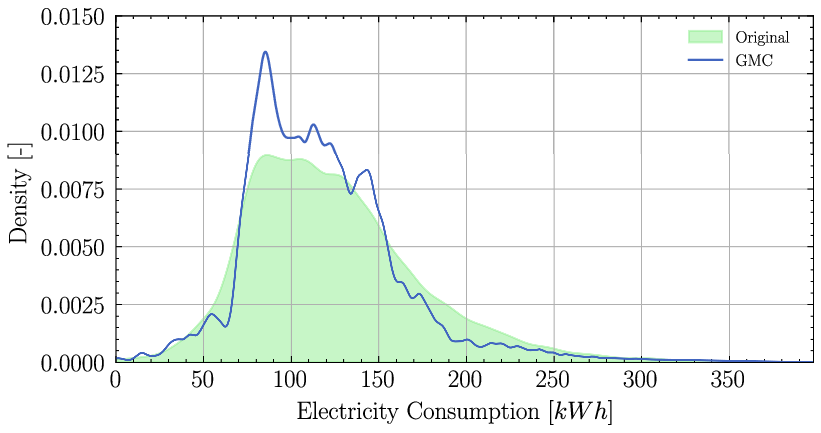}
    \caption{Distribution of the original and generated data.}
    \label{fig_prob_distri}
\end{figure}

The original data and results generated by the GMC model are depicted in Fig.~\ref{fig_prob_distri}. The GMC model captured the original patterns of peaks and valleys and diverse scenarios between different nodes in the testing distribution network. For instance, in the original data, the daily consumption profiles at around noon are diverse, where some nodes equipped with ESSs have negative load consumption (discharged), while others show peaks of daily consumption. The developed GMC model replicates such diversity. 

\begin{figure}[ht]
    \centering
    \includegraphics[width=1.0\columnwidth]{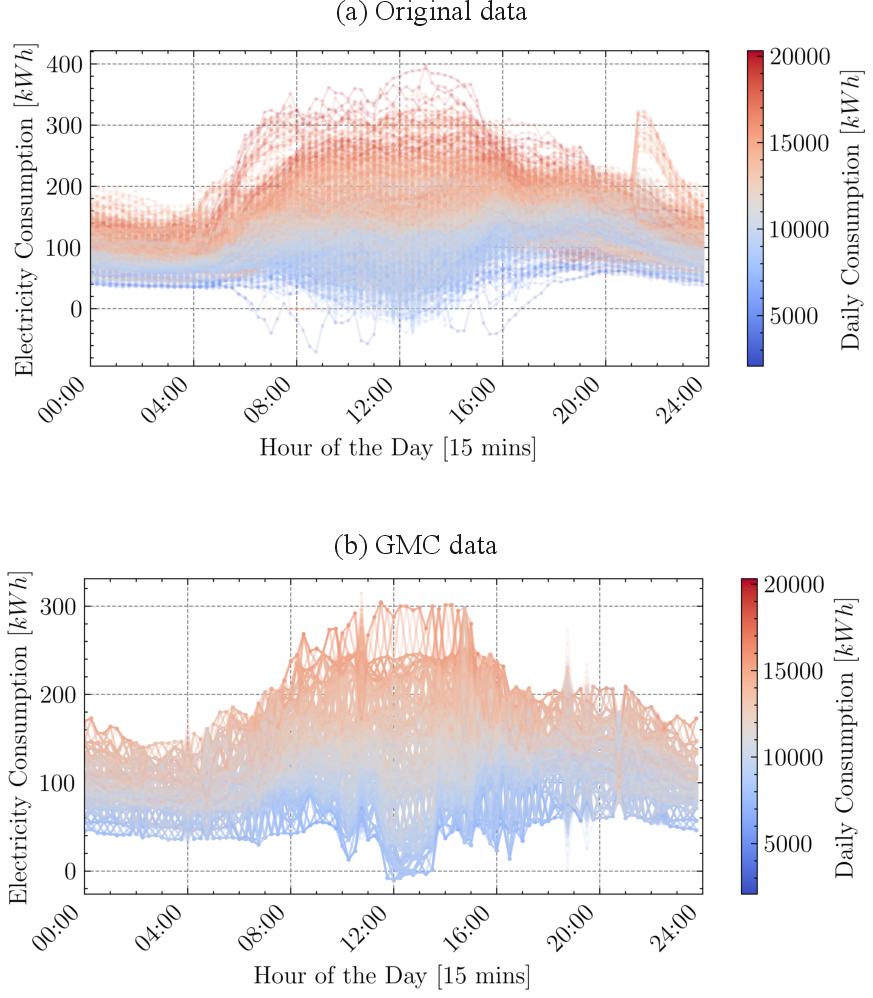}
    \caption{Original and GMC generated load profiles. The color of the profiles corresponds to the sum of daily consumption.}
    \label{fig_data_compare}
\end{figure}

Fig.~\ref{fig_data_compare} shows the original and generated data distribution shape. Both original and generated data have a long tail distribution. The shape of the GMC augmentation model's distribution matches the original data's shape. Therefore, the generated data profiles can enhance the scenario diversity without losing the original distribution and time correlation in the original dataset. 

\begin{table*}
\centering
\caption{Mean and 95\% confidence bounds for reward, violation penalty and performance bound.}
\label{quantitative analysis for performance on 30 days}
\scalebox{0.9}{
\begin{tblr}{
  cell{2}{1} = {r=3}{},
  cell{5}{1} = {r=3}{},
  hline{1,10} = {-}{0.08em},
  hline{2,5,8} = {-}{},
  hline{3-4,6-7,9} = {2-5}{dashed},
}
\textbf{Primary Dataset} & \textbf{Augmented Dataset}        & \textbf{Reward [-]}                                                           & \textbf{Violation Penlaty [-]}                             & \textbf{Performance bound $[\%]$}        \\
One month                & No augmentation                   & {DDPG (3.40$\mypm$0.86)\\ \textbf{PPO (5.91$\mypm$0.91)}\\SAC (4.825$\mypm$0.62)\\TD3 (3.49$\mypm$0.88)}    & {DDPG (0.0$\mypm$0.0)\\ \textcolor{red}{PPO (-0.002$\mypm$0.001)}\\SAC (0.0$\mypm$0.0)\\TD3 (0.0$\mypm$0.0)}           & {DDPG (51.1$\mypm$6.7)\\ \textcolor{blue}{PPO (69.1$\mypm$4.8)}\\SAC (62.5$\mypm$4.1)\\TD3 (52.4$\mypm$7.0)} \\
                         & augment 1 year    & {DDPG (9.55$\mypm$0.88)\\ \textbf{PPO (11.625$\mypm$0.92)}\\SAC (9.95$\mypm$0.63)\\TD3 (10.565$\mypm$0.91)} & { \textcolor{red}{DDPG (-1.05$\mypm$-0.77)}\\PPO (-0.039$\mypm$-0.01)\\SAC (-0.25$\mypm$-0.01)\\TD3 (-0.09$\mypm$-0.01)} & {DDPG (82.8$\mypm$1.1)\\ \textcolor{blue}{PPO (84.0$\mypm$1.0)}\\SAC (83.4$\mypm$0.5)\\TD3 (83.9$\mypm$0.9)} \\
                         & augment 5 year                    & {DDPG (7.37$\mypm$0.92)\\ \textbf{PPO (12.59$\mypm$0.88})\\SAC (8.25$\mypm$0.69)\\TD3 (8.02$\mypm$0.91)}                                         & {DDPG (-0.32$\mypm$-0.22)\\ \textcolor{red}{(PPO-2.10$\mypm$-0.69)}\\SAC (-0.18$\mypm$-0.09)\\TD3 (-0.96$\mypm$-0.41)}                      & {DDPG (76.35$\mypm$4.31)\\ \textcolor{blue}{PPO (85.9$\mypm$1.07)}\\SAC (79.58$\mypm$1.93)\\TD3 (78.82$\mypm$2.67)} \\
Three Month~             & No augmentation                   & {DDPG (8.54$\mypm$0.99)\\PPO (6.73$\mypm$0.97)\\SAC (6.92$\mypm$0.72)\\ \textbf{TD3 (8.60$\mypm$0.92)}}    & {DDPG (0.0$\mypm$0.0)\\PPO (0.0$\mypm$0.0)\\SAC (0.0$\mypm$0.0)\\TD3 (0.0$\mypm$0.0)}          & {DDPG (80.4$\mypm$2.3)\\PPO (73.5$\mypm$4.2)\\SAC (74.3$\mypm$3.1)\\ \textcolor{blue}{TD3 (80.6$\mypm$2.1)}} \\
                         & augment 1 year         & {DDPG (9.38$\mypm$0.99)\\ \textbf{PPO (9.68$\mypm$0.94)}\\SAC (7.78$\mypm$0.55)\\TD3 (9.24$\mypm$0.92)}    & {DDPG (0.0$\mypm$0.0)\\PPO (0.0$\mypm$0.0)\\SAC (0.0$\mypm$0.0)\\TD3 (0.0$\mypm$0.0)}          & {DDPG (82.5$\mypm$1.4)\\ \textcolor{blue}{PPO (83.0$\mypm$1.0)}\\SAC (78.0$\mypm$1.9)\\TD3 (82.2$\mypm$1.4)} \\
                         & augment 5 year  & {\textbf{DDPG (9.24$\mypm$0.89)}\\PPO (8.72$\mypm$0.97)\\SAC (6.02$\mypm$0.71)\\TD3 (8.45$\mypm$0.95)}                                         & {DDPG (0.0$\mypm$0.0)\\PPO (0.0$\mypm$0.0)\\SAC (0.0$\mypm$0.0)\\TD3 (0.0$\mypm$0.0)}                      & {\textcolor{blue}{DDPG (82.19$\mypm$1.4)}\\PPO (81.01$\mypm$3.1)\\SAC (69.71$\mypm$3.75)\\TD3 (80.20$\mypm$3.32)} \\
One year~                & No augmentation                   & {DDPG (7.061$\mypm$0.93)\\ \textbf{PPO (8.173$\mypm$1.02)}\\SAC (7.302$\mypm$0.84)\\TD3 (7.325$\mypm$1.03)} &  { \textcolor{red}{DDPG (-0.01$\mypm$0.0)}\\PPO (0.0$\mypm$0.0)\\SAC (0.0$\mypm$0.0)\\TD3 (0.0$\mypm$0.0)}        & {DDPG (75.0$\mypm$3.7)\\ \textcolor{blue}{PPO (79.3$\mypm$2.8)}\\SAC (76.1$\mypm$3.2)\\TD3 (76.2$\mypm$3.8)} \\
                         & augment 5 year                    & {DDPG (7.58$\mypm$0.79)\\\textbf{PPO (8.91$\mypm$0.87)}\\SAC (8.47$\mypm$0.86)\\TD3 (7.99$\mypm$0.99)}                                         & {DDPG (0.0$\mypm$0.0)\\PPO (0.0$\mypm$0.0)\\SAC (0.0$\mypm$0.0)\\TD3 (0.0$\mypm$0.0)}                      & {DDPG (77.20$\mypm$2.76)\\ \textcolor{blue}{PPO (81.44$\mypm$1.71)}\\ SAC (80.26$\mypm$2.12)\\TD3 (78.72$\mypm$2.90)} 
\end{tblr}}
\end{table*}

Table~\ref{quantitative analysis for performance on 30 days} presents the average reward, voltage magnitude violation penalty, and performance bounds for DRL algorithms on a separate 30-day test dataset. These algorithms, trained on primary datasets of 1 month, 3 months, and 1 year, were further augmented to 1 year and 5 years to examine the effects of data augmentation within the RL-ADN framework. Consistency in training parameters was maintained across 1000 episodes, and the results include 95\% confidence intervals.

Initially, the performance of DRL algorithms using 1-month data was suboptimal. For example, the PPO algorithm's highest performance bound was below 70\% (69.1\%). However, post-augmentation, there was a significant improvement: PPO's performance increased to 84.0\% and 85.9\% with 1-year and 5-year data augmentation, respectively. When trained on 3-month primary data, DRL algorithms demonstrated good performance, which was further enhanced with data augmentation. For instance, TD3 improved from 80.6\% to 82.2\% with 1-year augmentation. Similarly, algorithms trained on one-year primary data showed good performance with minimal test set violations, and augmentation yielded incremental performance gains, as seen with PPO's increase from 79.3\% to 81.44\%. These results underscore the significance of data augmentation in enhancing the adaptation of DRL algorithms to varied market conditions, particularly for algorithms like DDPG and TD3. In scenarios with limited original datasets, the data augmentation module in the RL-ADN framework can substantially raise the performance ceiling of DRL algorithms.

However, a concerning observation was the increase in voltage magnitude violations in the 1-month data set trained algorithms post-augmentation, particularly notable with the 5-year augmentation. This could be attributed to the augmented data increasing scenario diversity but not altering the data distribution, as illustrated in Fig.~\ref{fig_data_compare}. In such cases, while DRL algorithms perform better within the existing data distribution, they may incur violations in extreme scenarios not encountered during training. Notably, algorithms trained on more diverse datasets (three-month and one-year) exhibited better control over voltage violations. This is likely because these datasets encompassed the extreme scenarios present in the test sets. Yet, when comparing performance, algorithms trained on the one-year dataset displayed a lower performance ceiling than those trained on the three-month dataset. This suggests that while the one-year data provides a more diverse training environment, leading to potentially better generalization, it also presents a slower learning curve due to its complexity.

Generally, results indicate that in scenarios with limited original datasets, the data augmentation module in the RL-ADN framework can substantially raise the performance ceiling of DRL algorithms. Moreover, the distribution of data and the diversity of scenarios significantly impact the performance of DRL algorithms. Scenario diversity raises the performance ceiling, while data distribution affects the training difficulty and performance in extreme scenarios. While augmentation improves overall performance, it introduces complexities like increased violation penalties, especially when the primary dataset has a limited data distribution.



\subsection{Enhancement of computation efficiency}

\begin{figure}[ht]
    \centering
    \includegraphics[width=1.0\columnwidth]{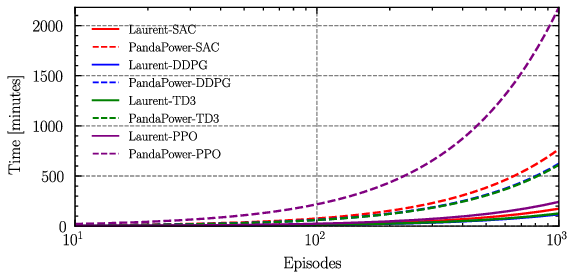}
    \caption{Training time for DRL algorithms with Laurent power flow and Panda power. The 34-node distribution network is used as a benchmark.}
    \label{fig_test_training_time}
\end{figure}

\begin{figure}[ht]
    \centering
    \includegraphics[width=1.0\columnwidth]{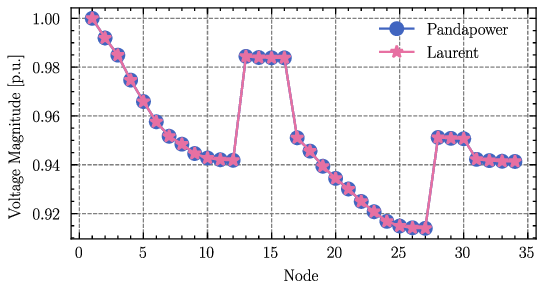}
    \caption{Voltage magnitude calculated by Laurent power flow and Panda power. The 34-node distribution network is used as a benchmark.}
    \label{fig_test_operation}
\end{figure}

\begin{table*}[ht]
\centering
\caption{Average calculation time comparison between Laurent power flow and PandaPower power flow for different scale distribution networks}
\label{tab_compare_time}
\scalebox{0.9}{
\begin{tabular}{c|cc|cc}
\toprule
\multirow{2}{*}{\textbf{ Distribution Networks}} & \multicolumn{2}{c|}{\textbf{Laurent Power}} & \multicolumn{2}{c}{\textbf{PandaPower}} \\
& Power Flow $[ms]$ & Env. Steps $[ms]$ & Power Flow $[ms]$ & Env. Steps $[ms]$ \\ 
\midrule
25 Nodes & \textbf{0.59} & 2.81 & 28.08 & 30.30 \\
\addlinespace
34 Nodes & \textbf{0.61} & 2.830 & 29.42 & 30.502 \\
\addlinespace
69 Nodes & \textbf{0.88} & 2.99 & 28.72 & 31.46 \\
\addlinespace
123 Nodes& \textbf{0.97} & 3.43 & 37.22 & 38.51 \\
\bottomrule
\end{tabular}}
\end{table*}

The performance comparison between Laurent power flow and PandaPower power flow was conducted across multiple scale distribution networks with node sizes: 25, 34, 69, and 123. The summarized results in Table \ref{tab_compare_time} indicate a distinct computational advantage for the Laurent power flow method over PandaPower.
First, Laurent power flow consistently maintained its efficiency, taking less than 1 ms across all node sizes. This is in stark contrast to PandaPower, which requires approximately 28 to 37 ms. In the smallest node size (25 nodes), Laurent is about 47 times faster than PandaPower when solving one-time power flow. As the node size grows to 123, the efficiency margin increases, with Laurent being nearly 38 times faster.

For executing one-time environment iteration, Laurent's time ranges from 2.8 to 3.4 ms, while PandaPower's duration extends from 30 to 38 ms. This indicates that, on average, Laurent is about ten times faster than PandaPower in processing environment steps, regardless of the node size.
Overall, the Laurent power flow displays a significant computational edge, particularly as the node size expands. This relative efficiency is pivotal in training DRL algorithms in large-scale distribution networks. The ability of the Laurent power flow to consistently outpace PandaPower across different node sizes underscores its scalability, making it a more versatile choice for varied applications.

The comparison between Laurent power flow and PandaPower flow algorithms across different DRL algorithms showcases significant time differences in training for the same number of episodes as shown in Fig.~\ref{fig_test_training_time}. A clear trend emerges from the data: the Laurent power flow consistently outperforms PandaPower in terms of computational efficiency. For the SAC algorithm, the Laurent power flow is approximately 4.4 times faster than the PandaPower flow. Similarly, for DDPG, the Laurent method shows a speedup of around 5.2 times. The TD3 algorithm with the Laurent technology is about 4.8 times faster. The most pronounced difference is observed in the PPO algorithm, where Laurent power flow is significantly faster, clocking at approximately 9.1 times the speed of PandaPower. PPO requires 2200 minutes for training,  making it the least efficient in this scenario. This is because PPO is an off-policy algorithm, which can not fully make use of the past experiences in the replay buffer, resulting in the lowest data efficiency and training speed.  On the other hand, DDPG emerges as the fastest, closely followed by TD3 and then SAC.

In conclusion, the Laurent power flow demonstrates a clear computational advantage across all tested algorithms. While the choice of algorithm also affects the training time, with PPO consistently taking the longest, the underlying power flow technology plays a crucial role in determining the overall efficiency. These findings can guide researchers and practitioners in making informed decisions when it comes to selecting the most efficient combination of power flow technology and reinforcement learning algorithm.

Fig.~\ref{fig_test_operation} displays the voltage magnitude results of a 34-node distribution network from Laurent power flow and PandaPower flow, respectively. The voltage magnitude results from both algorithms remain almost the same magnitude, with an average error of no more than 0.0001\%. Such a high precision from Laurent power flow can track the real voltage dynamics accurately, regrading of the load changes. Moreover, the integration of Laurent power flow with the developed environment can significantly save the time cost for a large magnitude power flow iteration during the training. Thus, our framework can accelerate notably training speed of DRL algorithms, without losing simulation precision.

\section{Conclusion}
This paper unveiled RL-ADN, an open-sourced library tailored for designing and implementing DRL environments for optimal ESS dispatch challenges in modern distribution networks. We highlighted the potential of advanced DRL algorithms, showcasing their capacity to yield near-optimal decisions. The first significant innovation of our approach is the seamless integration of the Laurent power flow, which offers unparalleled computational advantages over traditional methods, achieving over tenfold faster. Another innovation is that RL-ADN integrates the Gaussian mixture model and Copula functions to augment the training dataset, thus further improving the performance ceiling for DRL algorithms.  We believe RL-ADN presents a unique and extensive platform for future DRL research in energy systems. This research underscores the potential of a modular, customizable, and efficient RL environment in addressing the complexities of the energy landscape. We anticipate that RL-ADN will inspire a new wave of studies in the energy domain, leveraging its adaptability and precision.

\appendix
\section{Mathematical formulation of optimal ESS dispatch Tasks}
\label{sec_math_formulation}

The template energy arbitrage task can be formulated by using the nonlinear programming (NLP) formulation given by \eqref{eq:real_goal}--\eqref{eq_SOC_cons}. The objective function in ~\eqref{eq:goal} is extended to \eqref{eq:real_goal}, aiming to minimize the total operational cost over the time horizon~${\cal T}$, comprising the cost of importing power from the main grid. The operational cost $\rho_{t}$ at time slot $t$ is settled according to the balancing market prices $\rho_t$ in EUR/MWh.
\begin{equation}\label{eq:real_goal}
    \min_{\substack{P^{B}_{m,t}, \forall m \in {\cal B}, \forall t \in {\cal T}}} \left\lbrace  \sum_{t \in {\cal T}} \left[ \rho_{t}\sum_{m \in {\cal N}} \left(P^D_{m,t} + P^{B}_{m,t} - P^{PV}_{m,t}\right)\Delta t \right] \right\rbrace.
\end{equation}
Subject to:
%
\vspace{-2mm}
\begin{multline} \label{eq:active_power_balance}
 \hspace{-5mm} \sum_{nm \in {\cal L}} P_{nm,t} - \sum_{mn \in {\cal L}} (P_{mn,t} + R_{mn}I_{mn,t}^2) + P_{m,t}^{B} \\ + P_{m,t}^{PV}+ P_{m,t}^{S}= P_{m,t}^{D}  \quad \forall m \in {\cal N}, \forall t \in {\cal T}  
\end{multline}
\vspace{-6mm}
\begin{multline} \label{eq:reactive_power_balance}
 \hspace{-5mm} \sum_{nm \in {\cal L}} Q_{nm,t} - \sum_{mn \in {\cal L}} (Q_{mn,t} + X_{mn}I_{mn,t}^2) + Q_{m,t}^{S} = Q_{m,t}^{D}  \\  \forall m \in {\cal N}, \forall t \in {\cal T}
\end{multline}
\vspace{-6mm}
\begin{multline}
\label{eq_votlage_drop}
 \hspace{-4mm} V_{m,t}^2-V_{n,t}^2=2(R_{mn}P_{mn,t}+X_{mn}Q_{mn,t})+ \\(R_{mn}^2+X_{mn}^2)I_{mn,t}^2 \quad  \forall m,n \in {\cal N}, \forall t \in {\cal T}  
\end{multline}
\vspace{-6mm}
\begin{flalign}
& V_{m,t}^2I_{mn,t}^2=P_{mn,t}^2+Q_{mn,t}^2 & \forall m,n \in {\cal N}, \forall t \in {\cal T} \label{eq_vi=pq} &     
\end{flalign}
\vspace{-6mm}
\begin{multline}
\hspace{-4mm} SOC_{m,t}^{B}=SOC_{m,t-1}^{B} + \eta^{B}_{m}P_{m,t}^{B}\Delta t/\overline{E}^{B}_{m} \hspace{0.30em} \forall m \in {\cal{B}}, \forall t \in {\cal{T}} \hspace{-0.4em} \label{eq_SOC_cha}
\end{multline}
\vspace{-6mm}
\begin{flalign}
& \underline{SOC}_{m}^{B}\leq SOC_{m,t}^{B}\leq\overline{SOC}_{m}^{B} & \forall m \in {\cal{B}}, \forall t \in {\cal{T}} & \label{eq_SOC_cons}\\
& \underline{P}^{B}_{m}\leq P^{B}_{m,t}\leq \overline{P}^{B}_{m} & \forall m \in {\cal B}, \forall t \in {\cal T} & \label{eq_char_disc_cons}\\
& \underline{V}^{2}\leq V_{m,t}^2\leq \overline{V}^{2} & \forall m \in {\cal N}, \forall t \in {\cal T} & \label{eq:voltage_boundary}\\
& 0 \leq I_{mn,t}^2 \leq \overline{I}_{mn}^{2} & \forall mn \in {\cal L}, \forall t \in {\cal T} & \label{eq:limites_corre_5}\\
& P_{m,t}^{S} = Q_{m,t}^{S} = 0 & \forall m \in {\cal N} \backslash \{1\}, \forall t \in {\cal T} & \label{eq:power_not_substation}
\end{flalign}

The grid level constraints are modeled using the power flow formulation shown in \eqref{eq:active_power_balance}--\eqref{eq_vi=pq} in terms of the active $P_{mn,t}$ power, reactive power $Q_{mn,t}$ and current magnitude $I_{mn,t}$ of lines, and the voltage magnitude $V_{m,t}$ of nodes. \eqref{eq:voltage_boundary} and \eqref{eq:limites_corre_5} enforce the voltage magnitude and line current limits, respectively, while \eqref{eq:power_not_substation} enforces that only one node is connected to the substation.
The energy storage system constraints are modeled by~\eqref{eq_SOC_cha}--\eqref{eq_char_disc_cons}. Equation~\eqref{eq_SOC_cha} models the dynamics of the ESSs' SOC on the set ${\cal B}$, while \eqref{eq_SOC_cons} enforces the SOC limits. Hereafter, it is assumed that the ESS $m \in {\cal B}$ is connected to node $m$, thus, ${\cal B} \subseteq {\cal N}$. Finally, \eqref{eq_char_disc_cons} enforces the  ESSs discharge/charge operation limits,  
Notice that to solve the above-presented sequential decision problem, all long-term operational data (e.g., expected PV generation and consumption) must be collected to properly define the EESs' dispatch decisions, while the power flow formulation must also be considered to enforce the voltage and current magnitude limits.


\section{Workflows for modules in RL-ADN}
\subsection{Data manager workflow}
\label{sec_data_manager_workflow}
\texttt{GeneralPowerDataManager} modular, is a unified data manager. Designed for automation, this class standardize various data preprocessing tasks, as follows:
\begin{itemize}
\item Loads time-indexed data directly from standard CSV files.
\item Classifies columns pertaining to active and reactive power, renewable energy generation, and electricity pricing, autonomously.
\item Clean and check the data, filling in missing values, ensuring data continuity and integrity.
\item Segregates the dataset into distinct training and test sets based on temporal delineation.
\item Offers utility methods, such as \texttt{select-timeslot-data} and \texttt{select-day-data}, enabling precise data extraction tailored to the RL training needs.
\end{itemize}

When the \texttt{GeneralPowerDataManager} class is initialized, it undergoes a series of operations: it verifies the data's integrity, replaces any NaN values, and partitions the dataset into training and testing parts as required. These preliminary tasks ensure that data quality is maintained and provide ease of access and utilization for subsequent RL training processes.

\subsection{Data Augmentation workflow}
\label{sec_data_augmentation_workflow}
The augmentation process involves several sophisticated statistical techniques, outlined as follows:
\begin{itemize}
\item The \texttt{ActivePowerDataManager} class, a subclass of the \texttt{GeneralPowerDataManager}, preprocesses the input data, fills missing values through interpolation, and restructures the data into an appropriate format for augmentation.
\item A Gaussian Mixture Model (GMM) is fitted to the marginal distribution of historical active power data for each node and time step, capturing the underlying distribution of power consumption.
\item The Bayesian Information Criterion (BIC) is employed to select the optimal number of components for each GMM, ensuring that the model complexity is balanced against the goodness of fit.
\item A Copula-based approach is then applied, which models the dependency structure between different nodes and time steps, allowing for the generation of synthetic data points that maintain the correlation observed in historical data.
\item The \texttt{augment\_data} method leverages the GMM and Copula to produce new data samples, which are then transformed from the probabilistic space back to the power data scale.
\end{itemize}

The \texttt{TimeSeriesDataAugmentor} modular interacts with the data manager to retrieve the necessary preprocessed data, and then applies its augmentation algorithms to produce an augmented dataset. The output is a synthetic yet realistic dataset that reflects the variability and unpredictability inherent in power systems. This enriched dataset is crucial for training RL agents, providing them with a diverse range of scenarios to learn from and ultimately resulting in a more adaptable and robust decision-making policy.

Upon completion of the augmentation process, the synthetic data is saved to a CSV file, facilitating easy integration into the training pipeline. This automated and sophisticated data augmentation procedure enhances the RL-ADN framework's capability to train more effective and resilient RL agents for the distribution network ESSs operations.

\printcredits
\bibliographystyle{IEEEtran}
\bibliography{cas-refs}

\end{document}